\documentclass[twoside,11pt]{article}
\usepackage{amsmath}
\usepackage{jmlr2e}
\usepackage{url}
\jmlrheading{12}{2011}{2825-2830}{3/11; Revised 8/11}{10/11}{Fabian Pedregosa, Ga\"el Varoquaux, Alexandre Gramfort, Vincent Michel, Bertrand Thirion, Olivier Grisel, Mathieu Blondel et al.}

\ShortHeadings{Scikit-learn: Machine Learning in Python}{Pedregosa, Varoquaux, Gramfort et al.}
\firstpageno{2825}

\begin{document}

\title{Scikit-learn: Machine Learning in Python}

\author{\name Fabian Pedregosa \email fabian.pedregosa@inria.fr \\
        \name Ga\"el Varoquaux \email gael.varoquaux@normalesup.org  \\
        \name Alexandre Gramfort \email alexandre.gramfort@inria.fr \\
        \name Vincent Michel  \email vincent.michel@logilab.fr \\
        \name Bertrand Thirion  \email bertrand.thirion@inria.fr \\
        \addr Parietal, INRIA Saclay  \\Neurospin,
      B\^at 145, CEA Saclay \\Gif sur Yvette, France
        \AND
        \name Olivier Grisel \email olivier.grisel@ensta.fr \\
        \addr Nuxeo\\Paris, France
        \AND
        \name Mathieu Blondel \email mblondel@ai.cs.kobe-u.ac.jp \\
        \addr Kobe University  \\Kobe, Japan
        \AND
        \name Andreas M\"uller \email andreas.mueller@columbia.edu \\
        \addr Department of Computer Science \& Data Science Institute \\Columbia University \\New York, USA
        \AND
        \name Joel Nothman \email joel.nothman@gmail.com \\
        \addr Sydney Informatics Hub\\University of Sydney, NSW, Australia
        \AND
        \name Gilles Louppe \email g.louppe@ulg.ac.be \\
        \addr Dept. of EE \& CS\\University of Li\`ege \\Li\`ege, Belgium
        \AND
        \name Peter Prettenhofer \email peter.prettenhofer@gmail.com \\
        \addr Bauhaus-Universit\"at Weimar \\Weimar, Germany
        \AND
        \name Ron Weiss \email ronweiss@gmail.com \\
        \addr Google Inc \\New York, NY,  USA
        \AND
        \name Vincent Dubourg \email vincent.dubourg@gmail.com\\
        \addr Clermont Universit\'e, IFMA, EA 3867, LaMI\\
        Clermont-Ferrand, France
        \AND
        \name Jake Vanderplas \email vanderplas@astro.washington.edu\\
        \addr Astronomy Department \\University of Washington, Box 351580 \\
	Seattle, WA,  USA
	\AND
        \name Alexandre Passos \email alexandre.tp@gmail.com \\
        \addr IESL Lab \\UMass Amherst \\Amherst, USA
}

\editor{Mikio Braun}

\maketitle

\author{%
        \name David Cournapeau \email cournape@gmail.com \\
        \addr Enthought  \\Cambridge,
        UK
        \AND
        \name Matthieu Brucher \email matthieu.brucher@gmail.com \\
        \addr Total SA, CSTJF  \\Pau, France
        \AND
        \name Matthieu Perrot \email matthieu.perrot@cea.fr\\
        \name \'Edouard Duchesnay \email edouard.duchesnay@cea.fr \\
        \addr LNAO \\Neurospin,
      B\^at 145, CEA Saclay \\Gif sur Yvette -- France
}

\alttitle

\begin{abstract}%
\noindent\emph{Scikit-learn} is a Python module integrating a wide range of
state-of-the-art machine learning algorithms for medium-scale supervised
and unsupervised problems. This package focuses on bringing machine
learning to non-specialists using a general-purpose high-level language.
Emphasis is put on ease of use, performance, documentation, and API
consistency.
It has minimal dependencies and is distributed under the simplified BSD
license, encouraging its use in both academic and commercial settings.
Source code, binaries, and documentation can be downloaded from
\url{http://scikit-learn.org}.
\end{abstract}


\section{Introduction}

The Python programming language is establishing itself as one of the
most popular languages for scientific computing. Thanks to its
high-level interactive nature and its maturing ecosystem of scientific
libraries, it is an appealing choice for algorithmic development and
exploratory data analysis \citep{cise2007,cise2011}. Yet, as a
general-purpose language, it is increasingly used not only in academic
settings but also in industry.

{\sl Scikit-learn} harnesses this rich environment to provide state-of-the-art
implementations of
many well known machine learning algorithms, while maintaining an
easy-to-use interface tightly integrated with the Python language. This answers the
growing need for statistical data analysis by non-specialists in the software and web
industries, as well as in fields outside of computer-science, such as biology or physics.
\noindent\emph{Scikit-learn} differs from other machine learning toolboxes in Python for
various reasons: \emph{i)} it is distributed under the BSD license
\noindent\emph{ii)} it incorporates compiled code for efficiency, unlike MDP
\citep{zito2008} and pybrain \citep{schaul2010}, \emph{iii)} it depends
only on numpy and scipy to facilitate easy distribution,
unlike pymvpa \citep{hanke2009} that has optional dependencies such
as R and shogun, and \emph{iv)} it focuses on imperative
programming, unlike pybrain which uses a data-flow framework.
While the package is mostly written in Python, it incorporates the C++
libraries LibSVM \citep{chang2001} and LibLinear \citep{fan2008} that
provide reference implementations of SVMs and generalized linear models
with compatible licenses.
Binary packages are available on a rich set of platforms including
Windows and any POSIX platforms. Furthermore, thanks to its liberal
license, it has been widely distributed as part
of major free software distributions such as Ubuntu, Debian, Mandriva,
NetBSD and Macports and in commercial distributions such as the ``Enthought
Python Distribution''.

\section{Project Vision}

\noindent\emph{Code quality.}
Rather than providing as many features as possible, the project's goal has been to provide solid
implementations. Code quality is ensured with unit
tests---as of release 0.8, test coverage is 81\%---and the use of static
analysis tools such as {\tt pyflakes} and {\tt pep8}. Finally, we
strive to use consistent naming for the functions and
parameters used throughout a strict adherence to the Python coding
guidelines and numpy style documentation.

\noindent\emph{BSD licensing.}
Most of the Python ecosystem is licensed with non-copyleft licenses. While
such policy is beneficial for adoption
of these tools by commercial
projects, it does impose some restrictions: we are unable to use some existing
scientific code, such as the GSL.

\noindent\emph{Bare-bone design and API.}
To lower the barrier of entry, we avoid framework code and keep the number
of different objects to a minimum, relying on numpy arrays for data
containers.

\noindent\emph{Community-driven development.}
We base our development on collaborative tools such as git, github and
public mailing lists. External contributions are welcome and
encouraged.

\noindent\emph{Documentation.}
%
\emph{Scikit-learn} provides a $\sim$300 page user guide including
narrative documentation, class references, a tutorial, installation
instructions, as well as more than 60 examples, some featuring
real-world applications. We try to minimize the use of
machine-learning jargon, while maintaining precision with
regards to the algorithms employed.

\section{Underlying Technologies}

\noindent\emph{Numpy:}
the base data structure used for data and model parameters.
Input data is
presented as numpy arrays, thus integrating seamlessly
with other scientific Python libraries. Numpy's view-based memory
model limits copies, even when binding with compiled code
\citep{Vanderwalt2011}. It also
provides basic arithmetic operations.

\noindent\emph{Scipy:}
efficient algorithms for linear algebra, sparse matrix representation,
special functions and basic statistical functions. {\sl Scipy} has
bindings for many Fortran-based standard numerical packages, such as
LAPACK. This is important for ease of installation and portability, as
providing libraries around Fortran code can prove challenging on
various platforms.

\noindent\emph{Cython:}
a language for combining C in Python. Cython makes it easy to
reach the performance of compiled languages with
Python-like syntax and high-level operations. It is also used to
bind compiled libraries, eliminating the boilerplate code
of Python/C extensions.

\section{Code Design}

\noindent\emph{Objects specified by interface, not by inheritance.}
To facilitate the use of external objects with \emph{scikit-learn},
inheritance is not enforced; instead, code
conventions provide a consistent interface.
The central object is an {\tt estimator}, that implements a
{\tt fit} method, accepting as arguments an input data array and,
optionally, an array of labels for supervised problems. Supervised estimators,
such as SVM classifiers, can implement a {\tt predict} method. Some estimators,
that we call {\tt transformers}, for example, PCA, implement a {\tt
transform} method, returning modified input data.
Estimators may also provide a {\tt score} method, which is an increasing
evaluation of goodness of fit: a log-likelihood, or a negated loss function.
The other important object is the \emph{cross-validation iterator},
which provides pairs of train and test indices to split input
data, 
for example K-fold,
leave one out, or
stratified cross-validation.

\noindent\emph{Model selection.}
\noindent\emph{Scikit-learn} can evaluate an estimator's performance or select
parameters using cross-validation, optionally distributing the
computation to several cores.
This is accomplished by wrapping an estimator in a {\tt GridSearchCV}
object, where the ``CV'' stands for ``cross-validated''.
During the call to {\tt fit}, it selects the parameters
on a specified parameter grid, maximizing a score
(the {\tt score} method of the underlying estimator). {\tt
predict}, {\tt score}, or {\tt transform} are then delegated to the tuned
estimator. This object can therefore be used transparently as any other
estimator.
Cross validation can be made more efficient for certain
estimators by exploiting specific properties, such as warm restarts
or regularization paths \citep{friedman2010}. This is supported through special
objects, such as the {\tt LassoCV}.
Finally, a {\tt Pipeline} object can
combine several {\tt transformers} and an estimator to create a
combined estimator to, for example, apply dimension reduction before
fitting. It behaves as a standard estimator, and
{\tt GridSearchCV} therefore tune the parameters of all steps.

\section{High-level yet Efficient: Some Trade Offs}

While \emph{scikit-learn} focuses on ease of use, and is
mostly written in a high level language, care has been taken to maximize
computational efficiency. In Table \ref{tab:comparisons}, we compare
computation time for a few algorithms implemented in the major machine
learning toolkits accessible in Python. We use the Madelon data
set \citep{Guyon2004}, 4400 instances and 500 attributes,
The data set is quite large, but small enough for most
algorithms to run.

\begin{table}[t]
{\small
\hspace*{.03\linewidth}%
\begin{tabular}{l c c c c c c}
\hline\hline 
 & scikit-learn & mlpy & pybrain & pymvpa &  mdp & shogun \\ [0.5ex]
\hline
Support Vector Classification & {\bf 5.2} & 9.47 & 17.5 & 11.52 & 40.48 & 5.63 \\
Lasso (LARS) & {\bf 1.17} & 105.3   & - &  37.35 & - & - \\
Elastic Net & {\bf 0.52} & 73.7 & -  &  1.44  & -  & - \\
k-Nearest Neighbors & 0.57 & 1.41 & - &  {\bf 0.56} & 0.58 & 1.36 \\
PCA (9 components) & {\bf 0.18} & - & - & 8.93  & 0.47 & 0.33 \\
k-Means (9 clusters) & 1.34 &  0.79 & $\star$ & -  & 35.75 & {\bf 0.68} \\
License &  BSD & GPL & BSD  &  BSD  & BSD  & GPL \\
\hline
\end{tabular}

-: Not implemented. \hfill
$\star$: Does not converge within 1 hour.
}

\vspace*{-1.5ex}
\caption{
Time in seconds on the Madelon data set for various machine learning libraries exposed in Python:
MLPy \citep{albanese2008}, PyBrain \citep{schaul2010}, pymvpa
\citep{hanke2009}, MDP \citep{zito2008} and Shogun
\citep{sonnenburg2010}.
For more benchmarks see {\tt http://github.com/scikit-learn}.
\vspace*{-1.5em}\label{tab:comparisons}
}
\end{table}

\noindent\emph{SVM.}
While all of the packages compared call libsvm in the
background, the
performance of \emph{scikit-learn} can be explained by two factors.
First, our bindings avoid memory copies and have up to 40\% less overhead than
the original libsvm Python bindings.
Second, we patch libsvm to improve efficiency
on dense data, use a smaller memory footprint, and better use memory
alignment and pipelining capabilities of modern processors. This patched
version also provides unique features, such as
setting weights for individual samples.

\noindent\emph{LARS.}
Iteratively refining the residuals instead of recomputing them gives
performance gains of 2--10 times over the reference R implementation
\citep{LARS}. {\sl Pymvpa} uses this implementation via the Rpy R
bindings and pays a heavy price to memory copies.

\noindent\emph{Elastic Net.}
We benchmarked the \emph{scikit-learn} coordinate descent implementations of
Elastic Net. It achieves the same order of
performance as the highly optimized Fortran version \emph{glmnet}
\citep{friedman2010} on medium-scale problems, but performance on very
large problems is limited since we do not use the KKT conditions to
define an active set.

\noindent\emph{kNN.}
The k-nearest neighbors classifier implementation constructs a ball
tree \citep{omohundro1989} of the samples, but uses a more
efficient brute force search in large dimensions.

\noindent\emph{PCA.}
For medium to large data sets, \emph{scikit-learn} provides an
implementation of a truncated PCA based on random projections
\citep{rokhlin2009}. 

\noindent\emph{k-means.}
\noindent\emph{scikit-learn}'s k-means algorithm is implemented in pure
Python.  Its performance is limited by the fact that numpy's
array operations take multiple passes over data.

\section{Conclusion}

\emph{Scikit-learn} exposes a wide variety of machine learning
algorithms, both supervised and unsupervised, using a consistent,
task-oriented interface, thus enabling easy comparison of methods for a
given application.
Since it relies on the scientific Python ecosystem, it can easily be
integrated into applications outside the traditional range of statistical
data analysis. Importantly, the algorithms, implemented in a high-level
language, can be used as building blocks for approaches
specific to a use case, for example, in medical imaging \citep{Michel2011}.
Future work includes \emph{online} learning, to scale to
large data sets.

\bibliography{scikit}

\end{document}